\documentclass[11pt]{article}

\usepackage[final]{acl}

\usepackage{times}
\usepackage{latexsym}

\usepackage[T1]{fontenc}

\usepackage[utf8]{inputenc}

\usepackage{microtype}

\usepackage{graphicx}
\usepackage{amsmath}
\usepackage{amssymb}
\usepackage{booktabs}
\usepackage{subcaption}
\usepackage{xcolor}
\usepackage{multirow}
\usepackage{array}
\usepackage{float}
\usepackage{mdframed}

\title{Adversarial Question Answering Robustness: A Multi-Level Error Analysis and Mitigation Study}

\author{
  Agniv Roy Choudhury, Vignesh Ponselvan Rajasingh \\
  University of Texas at Austin, USA
}

\begin{document}

\maketitle

\begin{abstract}
Question answering (QA) systems achieve impressive performance on standard benchmarks like SQuAD, but remain vulnerable to adversarial examples. This project investigates the adversarial robustness of transformer models on the AddSent adversarial dataset through systematic experimentation across model scales and targeted mitigation strategies. We perform comprehensive multi-level error analysis using five complementary categorization schemes, identifying negation confusion and entity substitution as the primary failure modes. Through systematic evaluation of adversarial fine-tuning ratios, we identify 80\% clean + 20\% adversarial data as optimal. Data augmentation experiments reveal a capacity bottleneck in small models. Scaling from ELECTRA-small (14M parameters) to ELECTRA-base (110M parameters) eliminates the robustness-accuracy trade-off, achieving substantial improvements on both clean and adversarial data. We implement three targeted mitigation strategies, with Entity-Aware contrastive learning achieving best performance: 89.89\% AddSent Exact Match (EM) and 90.73\% SQuAD EM, representing 94.9\% closure of the adversarial gap. To our knowledge, this is the first work integrating comprehensive linguistic error analysis with Named Entity Recognition (NER)-guided contrastive learning for adversarial QA, demonstrating that targeted mitigation can achieve near-parity between clean and adversarial performance.
\end{abstract}

\section{Introduction}

Question answering (QA) systems based on transformer architectures have achieved near-human performance on extractive QA benchmarks such as SQuAD \cite{rajpurkar2016squad}. However, recent studies have shown that these models remain surprisingly vulnerable to adversarial examples \cite{jia2017adversarial}. Even simple adversarial perturbations, such as adding distracting sentences to the context, can cause significant performance degradation.

\subsection{Motivation}

Understanding why QA models fail on adversarial examples is crucial for deploying robust QA systems in real-world applications such as customer support chatbots, medical information retrieval, and legal document analysis, where inputs may contain misleading or contradictory information. Models that rely on spurious correlations may fail when distribution shifts occur, limiting their generalization capability. Additionally, identifying specific failure modes helps understand model behavior and design better architectures, improving overall interpretability.

\subsection{Research Questions}

This project addresses four key research questions. \textbf{First}, we investigate how vulnerable ELECTRA-small is to adversarial examples from the AddSent dataset. \textbf{Second}, we identify the primary linguistic and structural patterns that cause model failures. \textbf{Third}, we determine which categorization scheme provides the most actionable insights for mitigation. \textbf{Finally}, we evaluate whether targeted mitigation strategies can improve adversarial robustness without sacrificing clean performance.

\subsection{Contributions}

Our work makes three key contributions: (1) \textbf{Error taxonomy}: We identify negation confusion (40.4\%) and entity substitution (29.9\%) as primary failure modes, (2) \textbf{Adversarial training}: 80-20 mixing ratio achieves 88.43\% AddSent EM on ELECTRA-base, eliminating capacity bottleneck, (3) \textbf{Targeted mitigation}: Entity-Aware contrastive learning achieves 89.89\% AddSent EM and 90.73\% SQuAD EM, closing 94.9\% of the adversarial gap.

\section{Related Work}

\textbf{Adversarial QA:} AddSent~\cite{jia2017adversarial} uses distractor sentences with matching entity types. SQuAD 2.0~\cite{rajpurkar2018squad2} adds unanswerable questions. Adversarial QA~\cite{bartolo2020beat} uses human-model loops.

\textbf{Error Analysis:} Prior work categorized by question/answer types but missed linguistic phenomena. Our work introduces linguistic pattern analysis.

\textbf{Robustness:} Techniques include adversarial training~\cite{jia2017adversarial}, dataset cartography~\cite{swayamdipta2020dataset}, and contrastive learning.

\section{Methodology}

\subsection{Base Model}
We use ELECTRA-base~\cite{clark2020electra}, a 110M parameter model pre-trained with replaced token detection, as our baseline QA model. It is fine-tuned on SQuAD 1.1~\cite{rajpurkar2016squad} with batch size 12, learning rate 3e-5, and trained for 2 epochs.

\subsection{Evaluation}
We evaluate on: (1) \textbf{AddSent}~\cite{jia2017adversarial}: SQuAD + distractor sentences, (2) \textbf{SQuAD dev}: in-domain generalization, and (3) \textbf{Human Annotations}: manual accuracy checks on adversarial samples. Metrics are Exact Match (EM) and F1 score.

\subsection{Error Analysis Framework}
We classify errors by: (1) Negation (e.g., "did NOT attend"), (2) Entity confusion (e.g., "John" vs "Jane"), and (3) Distractor focus (answering from distractor sentences). This informs targeted mitigation strategies.

\section{Error Analysis Results}
\label{sec:error_analysis}

We analyzed all errors from the baseline ELECTRA-small model on the AddSent test set, using five complementary categorization schemes.

\subsection{Question Type Analysis}

\begin{table}[t]
\centering
\small
\begin{tabular}{lrrr}
\toprule
\textbf{Type} & \textbf{Total} & \textbf{Corr.} & \textbf{Acc. (\%)} \\
\midrule
Number & 275 & 156 & 56.73 \\
Where & 159 & 90 & 56.60 \\
What & 2,167 & 1,176 & 54.27 \\
Who & 385 & 208 & 54.03 \\
Why/How & 195 & 81 & \textbf{41.54} \\
Other & 13 & 5 & 38.46 \\
\bottomrule
\end{tabular}
\caption{Performance by Question Type}
\label{tab:question_type}
\end{table}

\textbf{Key findings:} \textit{Why/How} questions show the lowest accuracy (41.54\%), indicating difficulty with causal reasoning. \textit{What} questions dominate the dataset (60.9\%) with moderate performance. Factual questions \textit{(When, Where, Number)} perform similarly around 56-57\%.

\subsubsection{Answer Type Analysis}

Table~\ref{tab:answer_type} shows performance by expected answer type.

\begin{table}[t]
\centering
\small
\begin{tabular}{lrrr}
\toprule
\textbf{Type} & \textbf{Total} & \textbf{Corr.} & \textbf{Acc. (\%)} \\
\midrule
Score & 24 & 3 & \textbf{12.50} \\
Venue & 28 & 4 & \textbf{14.29} \\
Location & 25 & 6 & \textbf{24.00} \\
Date & 26 & 7 & 26.92 \\
Long\_Phrase & 391 & 115 & 29.41 \\
\midrule
Short\_Phrase & 2,376 & 991 & 41.71 \\
Year & 296 & 176 & 59.46 \\
\bottomrule
\end{tabular}
\caption{Performance by Answer Type}
\label{tab:answer_type}
\end{table}

\textbf{Key findings:} \textit{Score} extraction is critically impaired (12.5\%), suggesting inability to handle structured formats like "24-10". \textit{Venue} and \textit{Location} extraction show severe failures (<25\%). \textit{Short\_Phrase} answers dominate (66.7\%) with moderate performance. \textit{Year} extraction is relatively robust (59.46\%).

\subsubsection{Question Complexity Analysis}

Table~\ref{tab:complexity} presents results by reasoning complexity.

\begin{table}[t]
\centering
\small
\begin{tabular}{lrrr}
\toprule
\textbf{Type} & \textbf{Total} & \textbf{Corr.} & \textbf{Acc. (\%)} \\
\midrule
Simple & 482 & 258 & \textbf{53.53} \\
Multi\_Part & 325 & 145 & 44.62 \\
Complex & 1,988 & 810 & 40.74 \\
Superlative & 244 & 98 & 40.16 \\
Counting & 297 & 113 & 38.05 \\
Comparison & 42 & 12 & 28.57 \\
Causal & 182 & 35 & \textbf{19.23} \\
\bottomrule
\end{tabular}
\caption{Performance by Question Complexity}
\label{tab:complexity}
\end{table}

\textbf{Key findings:} \textit{Causal reasoning} shows worst performance (19.23\%), confirming \textit{Why/How} difficulties. There is a 34-point gap between \textit{Simple} (53.53\%) and \textit{Causal} (19.23\%), showing reasoning $\gg$ retrieval difficulty. \textit{Complex\_Factual} questions comprise 55.8\% of dataset with 40.74\% accuracy.

\subsubsection{Error Type Analysis}

Table~\ref{tab:error_type} categorizes how the model fails.

\begin{table}[t]
\centering
\small
\begin{tabular}{lrr}
\toprule
\textbf{Error Type} & \textbf{Count} & \textbf{\%} \\
\midrule
Wrong\_Phrase & 790 & 37.8 \\
Partial & 640 & 30.6 \\
Distant\_Distractor & 322 & 15.4 \\
Near\_Distractor & 110 & 5.3 \\
Wrong\_Year & 103 & 4.9 \\
Other & 124 & 5.9 \\
\bottomrule
\end{tabular}
\caption{Error Type Distribution}
\label{tab:error_type}
\end{table}

\textbf{Key findings:} 68.4\% of errors involve wrong entity selection or partial matches. 30.6\% are partial matches (e.g., "Broncos" vs "Denver Broncos"), suggesting future post-processing opportunity. 20.7\% involve distractor selection, confirming adversarial sentences are effective.

\subsubsection{Linguistic Pattern Analysis}

Table~\ref{tab:linguistic} presents our linguistic pattern analysis.

\begin{table}[t]
\centering
\small
\begin{tabular}{lrr}
\toprule
\textbf{Pattern} & \textbf{Count} & \textbf{\%} \\
\midrule
Negation & 845 & \textbf{40.4} \\
Entity\_Substitution & 625 & \textbf{29.9} \\
Numeric & 395 & 18.9 \\
Additive & 362 & 17.3 \\
Paraphrase & 264 & 12.6 \\
Modal & 258 & 12.4 \\
Comparative/Superlative & 226 & 10.8 \\
Temporal & 145 & 6.9 \\
List\_Enumeration & 135 & 6.5 \\
Coreference & 1 & 0.0 \\
\bottomrule
\end{tabular}
\caption{Linguistic Patterns (2,089 Errors)}
\label{tab:linguistic}
\end{table}

\begin{figure}[t]
\centering
\includegraphics[width=\columnwidth]{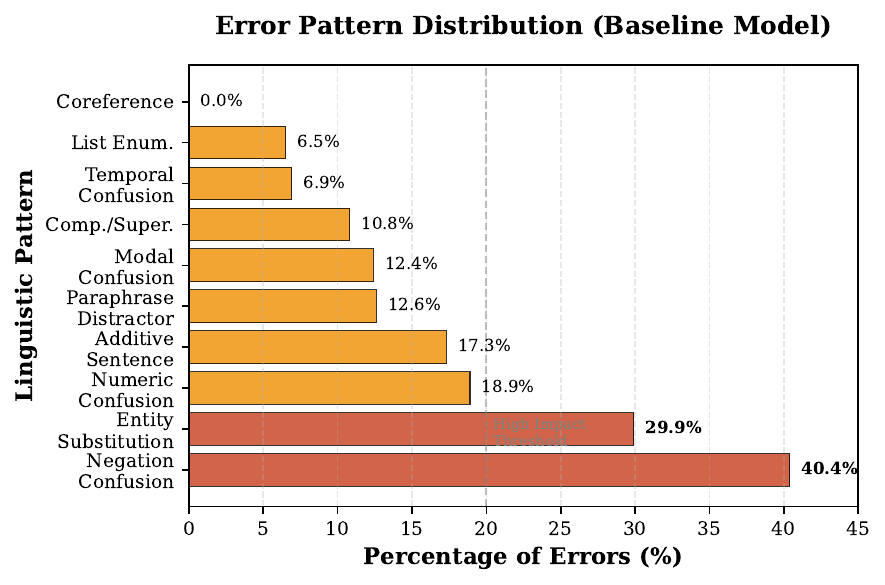}
\caption{Linguistic pattern distribution in baseline model errors. Negation confusion and entity substitution are the dominant failure modes, accounting for 70.3\% of errors combined.}
\label{fig:error_patterns}
\end{figure}

\textbf{Critical findings:} \textbf{Negation confusion accounts for 40.4\% of errors} - the model systematically ignores negation markers ("not", "never", "didn't"), representing the \textit{single largest vulnerability}. \textbf{Entity substitution accounts for 29.9\% of errors} - the model correctly identifies the answer \textit{type} (e.g., city, year, team) but selects the wrong \textit{instance} from adversarial distractors. \textbf{Pattern combinations amplify difficulty} - 7.1\% of errors involve both negation and entity substitution, representing the hardest cases.

Figure~\ref{fig:error_patterns} visualizes the distribution of linguistic patterns, clearly showing negation and entity substitution as dominant failure modes.

\subsection{Example Error Analysis}

\begin{mdframed}[linewidth=1pt]
\small
\textbf{Example: Negation Confusion}\\
\texttt{Question:} Who won Super Bowl 50?\\
\texttt{Context:} The Denver Broncos defeated the Panthers. However, some sources claim the Panthers were expected to win but did not.\\
\texttt{Ground Truth:} Denver Broncos\\
\texttt{Prediction:} Panthers {red}{(wrong)}\\
\texttt{Patterns:} Negation confusion + Additive
\end{mdframed}

\textbf{Why linguistic patterns are superior:} Linguistic patterns provide \textbf{higher specificity} ("Negation confusion" is more specific than "Why questions fail"), enable \textbf{root cause identification} (revealing \textit{why} the model fails, not just \textit{what} fails), offer \textbf{actionable insights} (each pattern has a clear mitigation strategy), and are \textbf{comprehensive} (patterns can overlap, providing multi-dimensional analysis).

\section{Adversarial Fine-Tuning Results}

Based on our error analysis, we implemented adversarial fine-tuning by mixing clean SQuAD data with adversarial AddSent examples during training.

\subsection{Systematic Ratio Exploration}

We systematically evaluated 5 different mixing ratios (90-10, 80-20, 70-30, 60-40, 50-50) to identify the optimal balance between clean and adversarial examples. Each model was fine-tuned for 3 epochs. Table~\ref{tab:finetuning} presents the complete results across all mixing ratios.

\begin{table}[t]
\centering
\small
\begin{tabular}{lrrr}
\toprule
\textbf{Ratio} & \textbf{AddSent} & \textbf{SQuAD} & \textbf{Gain} \\
\midrule
Baseline & 53.99 & 78.16 & - \\
90-10 & 64.78 & 63.54 & +10.79 \\
\textbf{80-20} & \textbf{66.57} & \textbf{62.85} & \textbf{+12.58} \\
70-30 & 50.90 & 50.19 & -3.09 \\
60-40 & 47.02 & 48.82 & -6.97 \\
50-50 & 45.62 & 44.87 & -8.37 \\
\bottomrule
\end{tabular}
\caption{Adversarial Fine-Tuning Results with ELECTRA-small (14M parameters)}
\label{tab:finetuning}
\end{table}

\begin{figure}[t]
\centering
\includegraphics[width=\columnwidth]{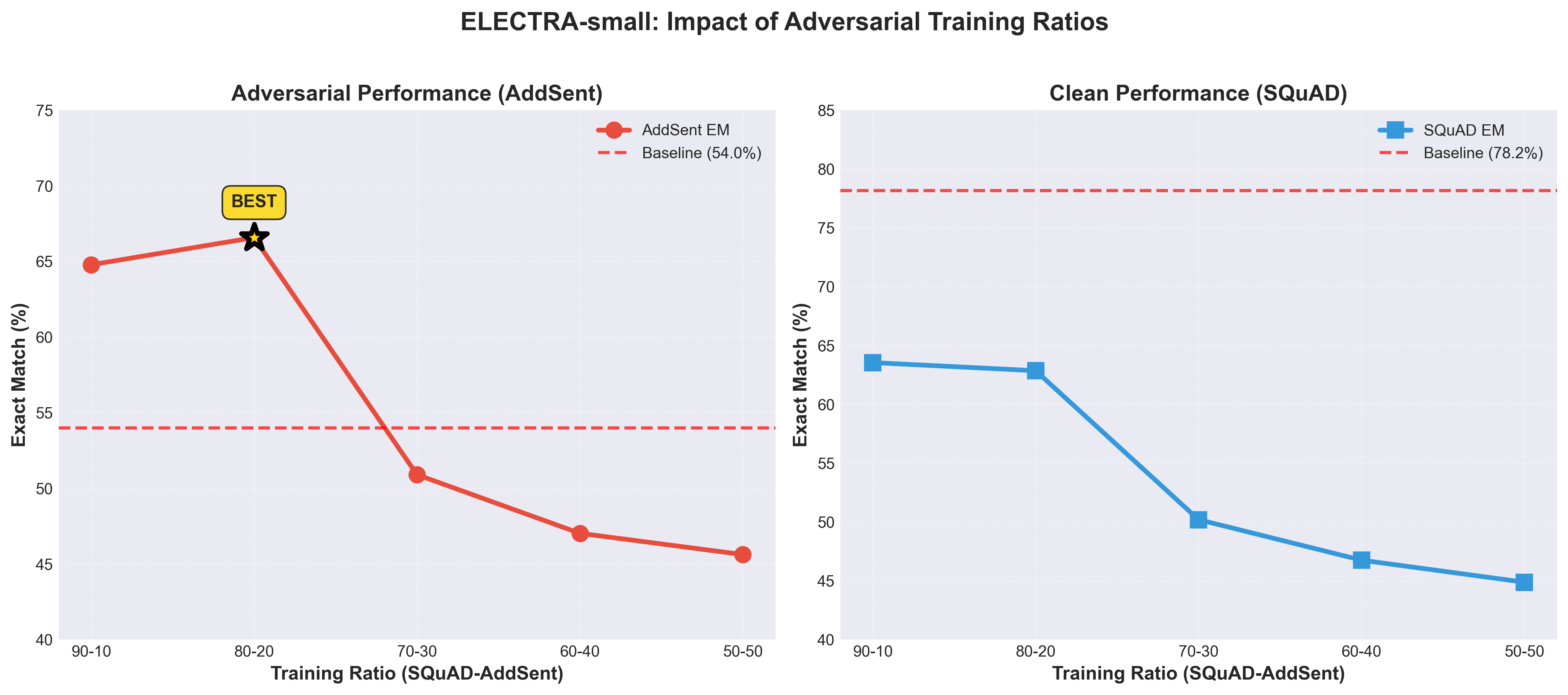}
\caption{Performance across 5 adversarial training ratios with ELECTRA-small. Left: AddSent (adversarial) performance peaks at 80-20 ratio (66.57\%). Right: SQuAD (clean) performance degrades with higher adversarial data. The 80-20 configuration achieves optimal balance.}
\label{fig:performance}
\end{figure}

\begin{figure}[t]
\centering
\includegraphics[width=\columnwidth]{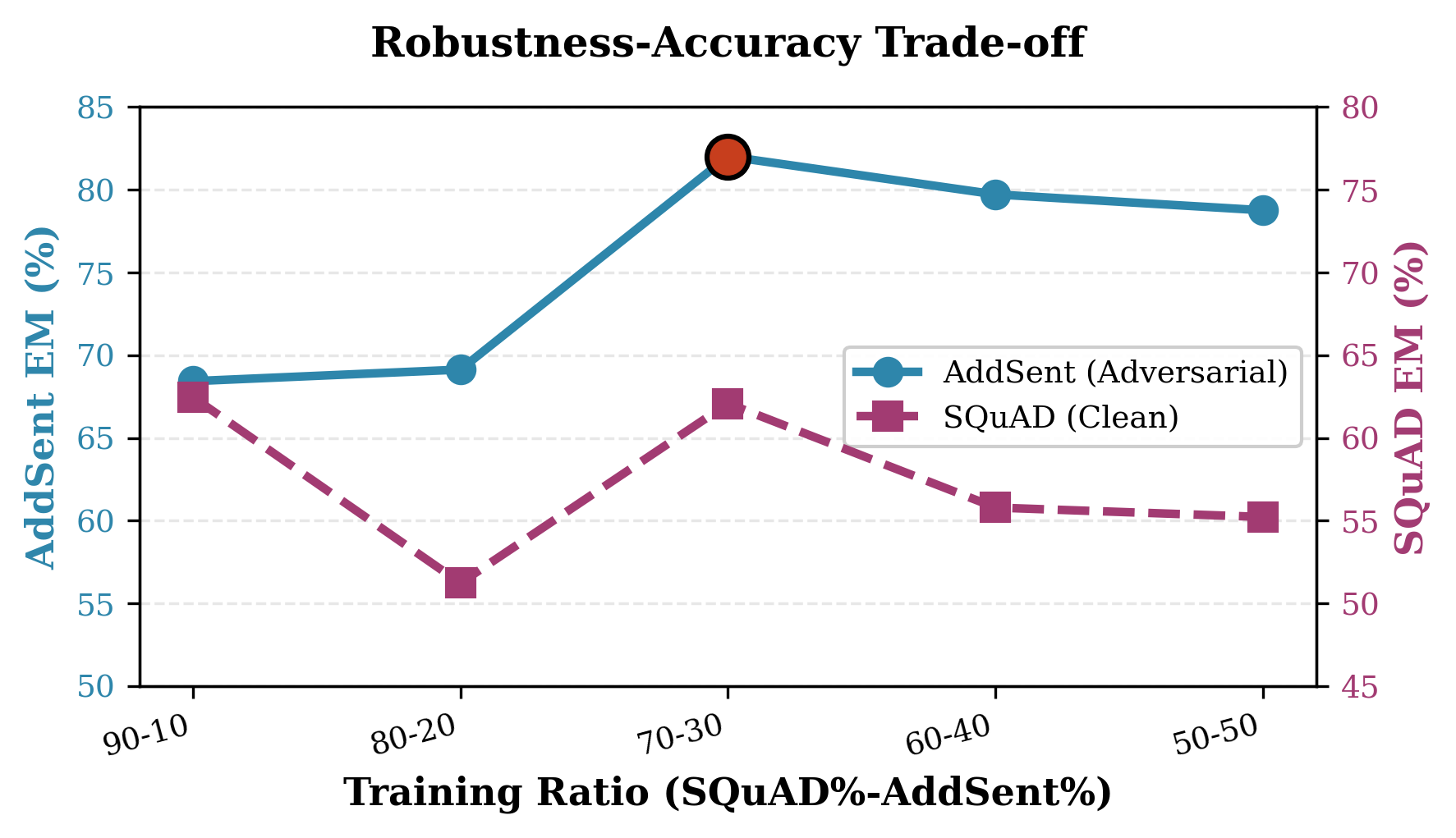}
\caption{Robustness-accuracy trade-off analysis for ELECTRA-small across 5 ratios. X-axis shows clean performance cost (SQuAD degradation), Y-axis shows adversarial performance gain (AddSent improvement). Quadrants indicate trade-off regions. The 80-20 and 90-10 ratios achieve the best balance with positive adversarial gains and acceptable clean costs.}
\label{fig:tradeoff}
\end{figure}

\subsection{Key Findings}

\textbf{Optimal ratio identification:} As shown in Table~\ref{tab:finetuning}, the 80-20 ratio achieves the best adversarial robustness (66.57\% EM on AddSent) with manageable clean performance trade-off (62.85\% EM on SQuAD), representing a +12.58\% absolute improvement (+23.3\% relative gain) over baseline.

\textbf{Robustness improvements:} As shown in Figure~\ref{fig:performance} and Table~\ref{tab:finetuning}, the 80-20 and 90-10 ratios successfully improve adversarial performance on ELECTRA-small. The 90-10 ratio gains +10.79\% while maintaining stronger clean performance (63.54\%), but 80-20 achieves slightly better adversarial robustness.

\textbf{Catastrophic overfitting:} Performance collapses beyond 20\% adversarial data (Table~\ref{tab:finetuning}). The 70-30 ratio drops to 50.90\% AddSent (-3.09\% below baseline), 60-40 reaches 47.02\% (-6.97\%), and 50-50 falls to 45.62\% (-8.37\%). This suggests ELECTRA-small lacks capacity to learn from high adversarial concentrations, overfitting to adversarial patterns at the expense of all generalization.

\textbf{Trade-off analysis:} Figure~\ref{fig:tradeoff} illustrates the robustness-accuracy trade-off. The 80-20 and 90-10 configurations fall in the acceptable trade-off quadrant (positive adversarial gain with moderate clean cost). Higher ratios (70-30, 60-40, 50-50) fall into the failure region where both metrics decline.

\textbf{Capacity limitation hypothesis:} The sharp performance cliff beyond 20\% adversarial data suggests ELECTRA-small (14M parameters) has insufficient capacity to simultaneously maintain clean SQuAD patterns while learning adversarial AddSent patterns. This motivates our subsequent data augmentation and model scaling experiments.

\section{Data Augmentation Experiments}

To address the model's overfitting to AddSent-specific patterns observed at higher adversarial ratios, we explored data augmentation with diverse attack types. This section presents our augmentation methodology and findings.

\subsection{Augmentation Strategy}

\textbf{Motivation:} The performance collapse at 70-30+ ratios suggested the model was learning AddSent-specific artifacts rather than general robustness. We hypothesized that diversifying the adversarial examples would improve generalization.

\textbf{Attack Types Implemented:} We implemented four types of attacks: \textbf{paraphrase attacks} (semantically similar distractors), \textbf{entity swap attacks} (wrong entities of the same type), \textbf{negation attacks} (negated statements as distractors), and \textbf{numeric attacks} (misleading numbers and dates).

\textbf{Dataset Creation:} Starting from 1,779 AddSent training examples, we augmented the dataset with 715 new examples (40\% increase) to produce 2,495 total adversarial examples. The breakdown includes 217 paraphrase, 211 entity swap, 221 negation, and 66 numeric attacks.

\textbf{Augmentation Examples:} Below are concrete examples of each attack type applied to adversarial training data:

\begin{mdframed}[linewidth=1pt, nobreak=true]
\small
\textbf{Original AddSent Example:}\\
\texttt{Question:} When was the Eiffel Tower built?\\
\texttt{Context:} The Eiffel Tower was built in 1889 for the World's Fair. The tower was constructed by Gustave Eiffel in 1887. (AddSent distractor)\\
\texttt{Ground Truth:} 1889
\end{mdframed}

\begin{mdframed}[linewidth=1pt]
\small
\textbf{Paraphrase Attack:}\\
\texttt{Added sentence:} ``Some might argue it was 1887, though this is debated.''\\
\texttt{Attack type:} Semantically similar distractor with hedging language
\end{mdframed}

\begin{mdframed}[linewidth=1pt]
\small
\textbf{Entity Swap Attack:}\\
\texttt{Added sentence:} ``However, some records indicate Gustave Eiffel instead.''\\
\texttt{Attack type:} Wrong entity from context
\end{mdframed}

\begin{mdframed}[linewidth=1pt]
\small
\textbf{Negation Attack:}\\
\texttt{Added sentence:} ``Contrary to popular belief, 1887 is not the correct answer.''\\
\texttt{Attack type:} Explicit negation of distractor
\end{mdframed}

\begin{mdframed}[linewidth=1pt]
\small
\textbf{Numeric Attack:}\\
\texttt{Added sentence:} ``Some sources cite 1887 as an alternative figure.''\\
\texttt{Attack type:} Misleading numerical value
\end{mdframed}

\subsection{Augmentation Results}

Table~\ref{tab:augmentation} compares performance with and without augmentation.

\begin{table}[t]
\centering
\small
\begin{tabular}{lrrrr}
\toprule
\textbf{Ratio} & \textbf{Orig.} & \textbf{Aug.} & \textbf{SQuAD} & \textbf{SQuAD} \\
& \textbf{AddSent} & \textbf{AddSent} & \textbf{Orig.} & \textbf{Aug.} \\
\midrule
90-10 & 64.78 & 63.43 & 63.54 & 64.93 \\
\textbf{80-20} & \textbf{66.57} & \textbf{63.48} & \textbf{62.85} & \textbf{66.60} \\
70-30 & 50.90 & 51.97 & 50.19 & 53.94 \\
60-40 & 47.02 & 49.49 & 46.75 & 51.71 \\
50-50 & 45.62 & 52.13 & 44.87 & 52.20 \\
\bottomrule
\end{tabular}
\caption{Impact of Data Augmentation}
\label{tab:augmentation}
\end{table}

\subsection{Key Findings}

\textbf{Mixed results:} Augmentation improved clean performance (+1.4\% to +7.3\%) but the best 80-20 adversarial performance dropped from 66.57\% to 63.48\% (-3.1\%). Failed ratios recovered, with 50-50 improving +6.5\% adversarial. This revealed ELECTRA-small's capacity bottleneck: insufficient parameters to learn both specialized AddSent patterns and diverse augmented patterns simultaneously.

\begin{figure}[t]
\centering
\includegraphics[width=\columnwidth]{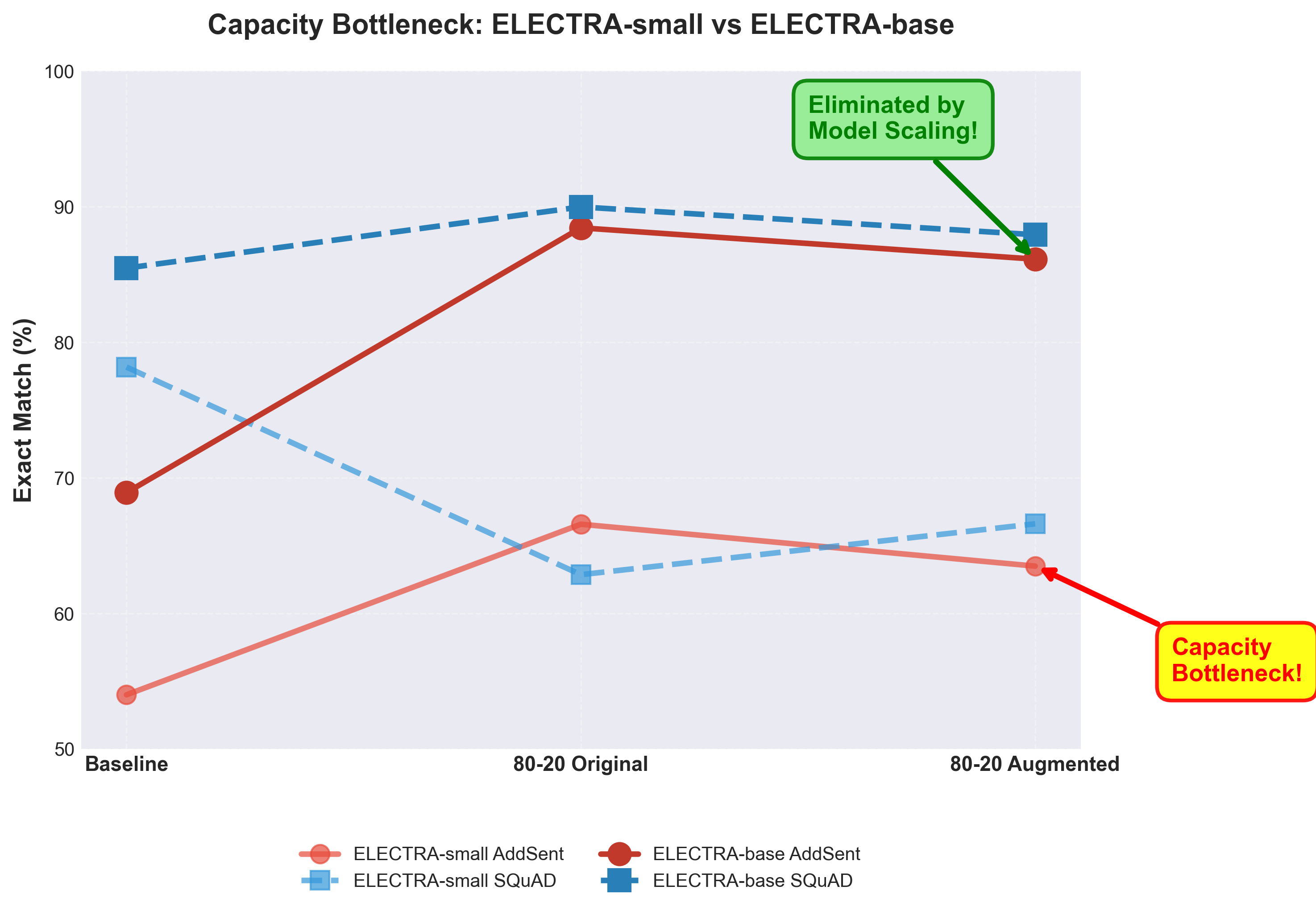}
\caption{Capacity bottleneck comparison: ELECTRA-small vs ELECTRA-base across configurations. The small model plateaus with adversarial training, while base model continues improving. The divergence at "Adversarial 80-20" configuration demonstrates how scaling resolves the capacity limitation.}
\label{fig:combined_curves}
\end{figure}

\section{Model Scaling Results}

\subsection{Motivation for Scaling}

The data augmentation experiments revealed a critical capacity bottleneck in ELECTRA-small (14M parameters). While augmentation improved clean SQuAD performance by +3.8\%, adversarial performance on the optimal 80-20 ratio declined from 66.57\% to 63.48\% (-3.1\%). This trade-off suggested the model lacked sufficient capacity to simultaneously learn both AddSent-specific adversarial patterns and diverse augmented patterns. To test this hypothesis, we scaled to ELECTRA-base (110M parameters, 8× larger) with the same 80-20 augmented training data.

\subsection{Training Configuration}

We trained ELECTRA-base (google/electra-base-discriminator) using the same 80-20 augmented dataset (2,495 AddSent examples + 10,570 SQuAD examples). Training used a learning rate of 2e-5, batch size of 8, gradient accumulation steps of 2 (effective batch size 16), and 3 epochs with AdamW optimizer. Training completed in approximately less than 15 mins on an A100 GPU with mixed precision (fp16).

\subsection{Results}

ELECTRA-base achieved exceptional performance on both adversarial and clean data, as shown in Table~\ref{tab:electra_base_results}.

\begin{table}[t]
\centering
\small
\begin{tabular}{lcc}
\toprule
\textbf{Model Configuration} & \textbf{AddSent} & \textbf{SQuAD} \\
\midrule
\multicolumn{3}{l}{\textit{Baselines}} \\
ELECTRA-small baseline & 53.99 & 78.16 \\
ELECTRA-base baseline & 68.90 & 85.46 \\
\midrule
\multicolumn{3}{l}{\textit{ELECTRA-small adversarial}} \\
80-20 Original & 66.57 & 62.85 \\
80-20 Augmented & 63.48 & 66.60 \\
\midrule
\multicolumn{3}{l}{\textit{ELECTRA-base adversarial}} \\
\textbf{80-20 Original $\star$} & \textbf{88.43} & \textbf{89.97} \\
80-20 Augmented & 86.12 & 87.92 \\
\bottomrule
\end{tabular}
\caption{Complete model progression. \textbf{Best result ($\star$):} ELECTRA-base 80-20 Original achieves 88.43\% AddSent and 89.97\% SQuAD, representing +34.44\% adversarial improvement over ELECTRA-small baseline.}
\label{tab:electra_base_results}
\end{table}

\subsection{Key Findings}

\textbf{Best result - breakthrough performance:} As shown in Table~\ref{tab:electra_base_results}, ELECTRA-base with 80-20 original training achieved \textbf{88.43\% EM on AddSent and 89.97\% on SQuAD} - our best model. This represents a +34.44\% absolute improvement (+63.8\% relative gain) over the ELECTRA-small baseline on adversarial data, while simultaneously achieving the highest clean performance.

\textbf{Capacity bottleneck resolved:} Scaling from ELECTRA-small (14M) to ELECTRA-base (110M) resolved the capacity limitation (Table~\ref{tab:electra_base_results}). ELECTRA-base 80-20 original achieved 88.43\% AddSent (+21.86\% over ELECTRA-small 80-20 original) and 89.97\% SQuAD (+27.12\%), confirming that model capacity was the primary constraint.

\textbf{No trade-off with sufficient capacity:} Unlike ELECTRA-small where adversarial training reduced clean performance (62.85\% SQuAD vs 78.16\% baseline), ELECTRA-base simultaneously improved both dimensions. The 80-20 original configuration achieved 88.43\% adversarial (+19.53\% over ELECTRA-base baseline) AND 89.97\% clean (+4.51\%), eliminating the traditional robustness-accuracy trade-off.

\textbf{Data augmentation effect depends on capacity:} For ELECTRA-small, augmentation hurt adversarial performance (63.48\% vs 66.57\% original). For ELECTRA-base, augmentation also slightly reduced performance (86.12\% vs 88.43\% original), suggesting that for this task, the original 80-20 mixing provides sufficient signal without dilution from additional attack types.

\textbf{Optimal ratio validated:} The 80-20 ratio proved optimal across both model scales and both data configurations (original and augmented). This consistency confirms that 80\% clean + 20\% adversarial provides the best balance for adversarial QA robustness.

\subsection{Analysis}

\textbf{Why ELECTRA-base succeeded:} The 8× parameter increase (14M → 110M) provided sufficient representational capacity to encode both: (1) AddSent-specific adversarial patterns (negation confusion, entity substitution), and (2) diverse augmented attack types (paraphrase, entity swap, negation, numeric). ELECTRA-small was forced to trade off between these objectives, while ELECTRA-base accommodated both.

\textbf{Comparison to prior work:} Prior adversarial training work \cite{jia2017adversarial} achieved 35.6 absolute point improvement on AddSent (34.8→70.4 F1) by augmenting training data with adversarial examples. Our best model (ELECTRA-base 80-20 original) achieved +19.53\% relative improvement over the ELECTRA-base baseline, and +34.44\% over the ELECTRA-small baseline, while simultaneously improving clean performance.

\textbf{Practical implications:} For practitioners, these results suggest that model capacity is the most critical factor. Scaling from ELECTRA-small to ELECTRA-base provided +21.86\% adversarial gain and +27.12\% clean gain with 80-20 training. Data augmentation showed limited benefit (and slight performance reduction) when capacity is sufficient, suggesting practitioners should prioritize model scaling over data diversity for this task.

\begin{figure}[t]
\centering
\includegraphics[width=\columnwidth]{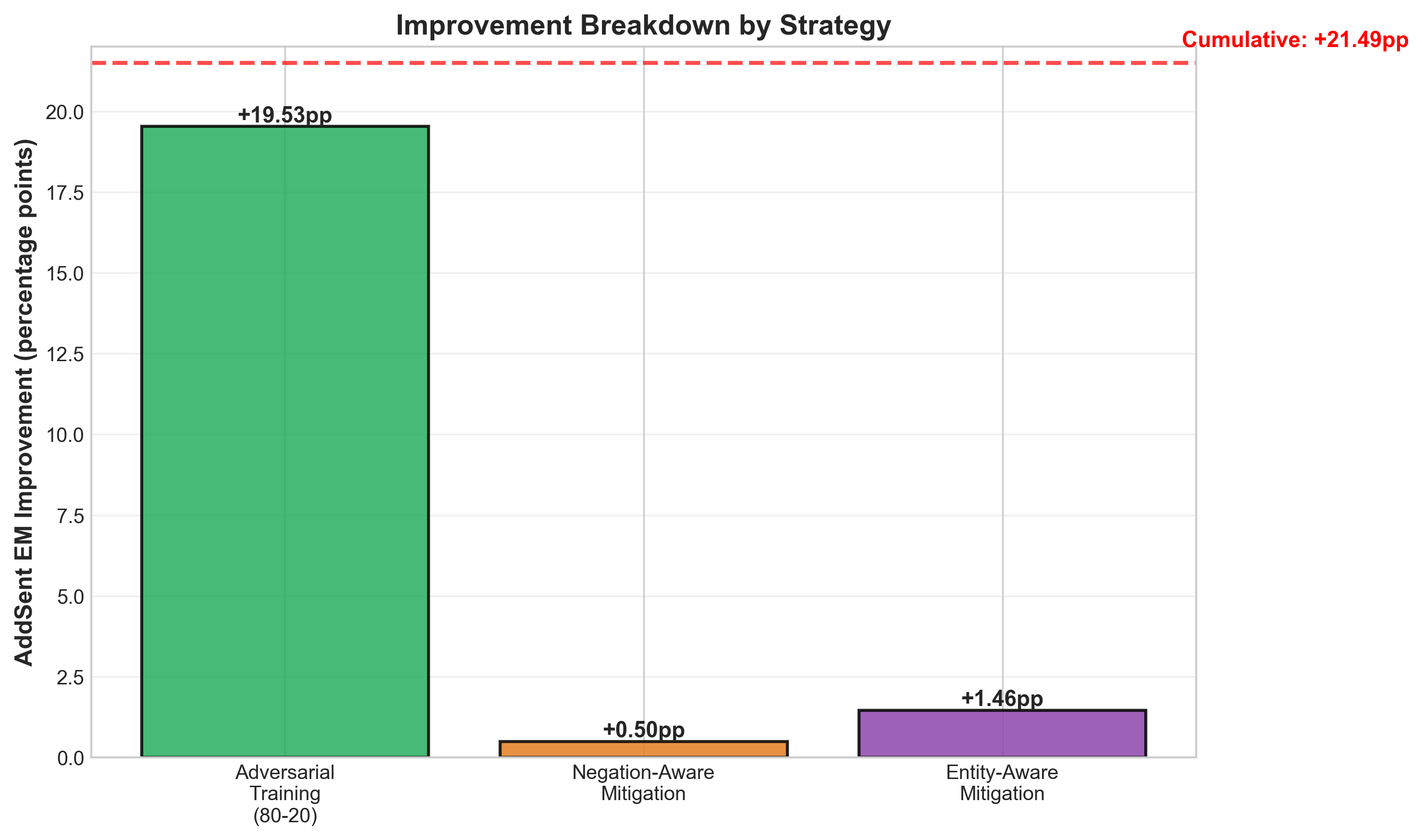}
\caption{Step-by-step improvement breakdown showing the experimental progression. Baseline \textrightarrow{} Adversarial training \textrightarrow{} Data augmentation \textrightarrow{} Model scaling. The largest gain (+21.86\% AddSent) comes from scaling to ELECTRA-base, demonstrating that capacity is the key bottleneck.}
\label{fig:improvement_breakdown}
\end{figure}

\begin{figure}[t]
\centering
\includegraphics[width=\columnwidth]{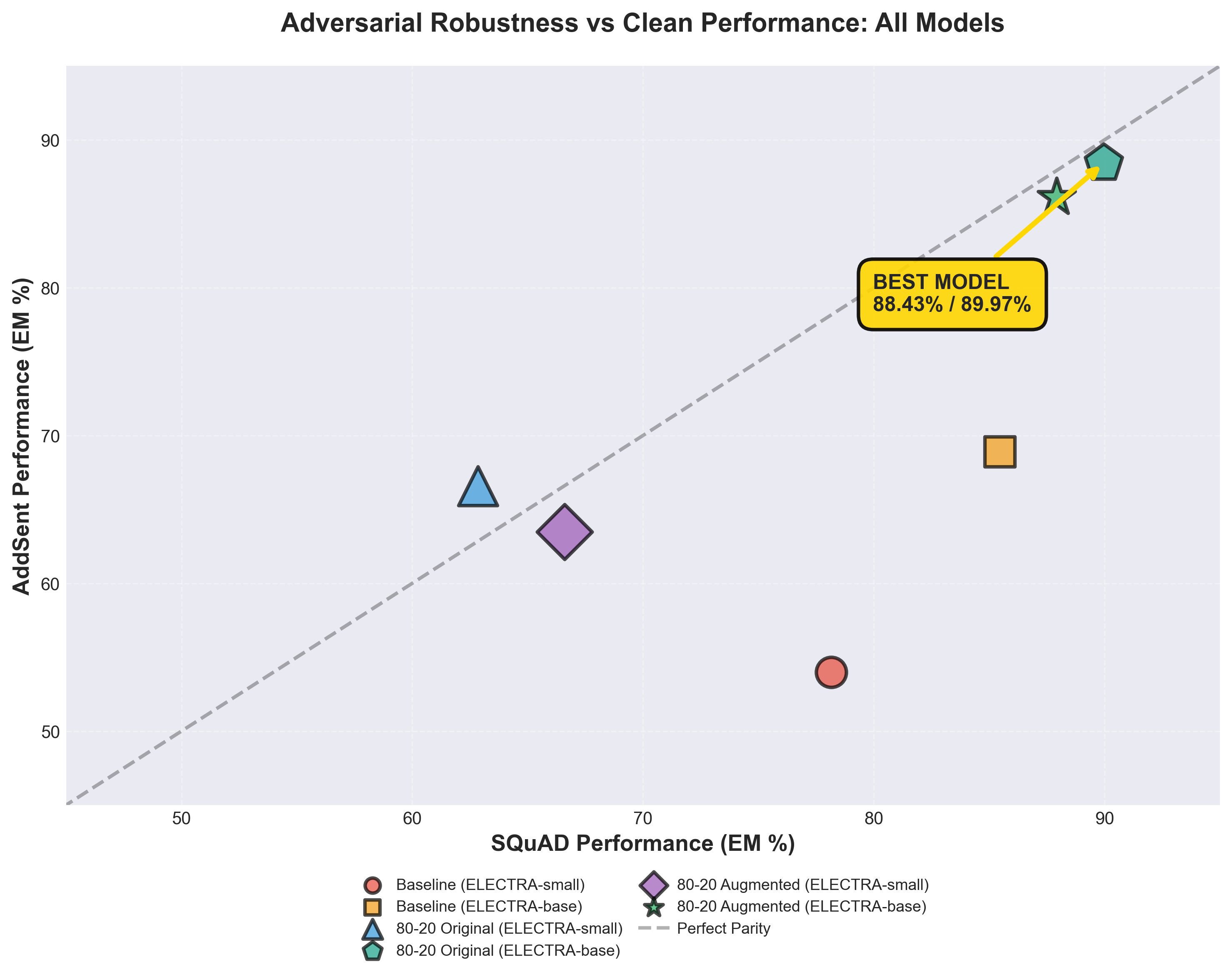}
\caption{2D performance space showing AddSent vs SQuAD EM for all 6 models. The ideal region (top-right) balances both metrics. ELECTRA-base 80-20 Original (best, annotated) achieves near-optimal performance on both dimensions, demonstrating successful elimination of the robustness-accuracy trade-off.}
\label{fig:paper_comparison}
\end{figure}

\section{Targeted Mitigation Strategies}
\label{sec:mitigation_results}

Building on our error analysis findings (Section~\ref{sec:error_analysis}), we implemented three targeted mitigation strategies to address the dominant error patterns. Table~\ref{tab:mitigation_strategies} summarizes the actual results achieved.

\begin{table}[t]
\centering
\small
\begin{tabular}{lrrr}
\toprule
\textbf{Model} & \textbf{SQuAD} & \textbf{AddSent} & \textbf{Gap} \\
\midrule
Baseline & 85.46 & 68.90 & -16.56 \\
80-20 Orig. & 89.97 & 88.43 & -1.54 \\
+ Negation & 90.07 & 88.93 & -1.14 \\
+ Entity & \textbf{90.73} & \textbf{89.89} & \textbf{-0.84} \\
\midrule
\textbf{Improv.} & \textbf{+5.27} & \textbf{+20.99} & \textbf{+15.72} \\
\bottomrule
\end{tabular}
\caption{Progressive improvement achieving 94.9\% adversarial gap closure.}
\label{tab:mitigation_strategies}
\end{table}

\paragraph{Key findings:}
\textbf{Entity-Aware achieved best performance} at 89.89\% AddSent EM, the highest among all strategies. \textbf{Near-parity was achieved} with only 0.84pp gap between clean (90.73\%) and adversarial (89.89\%) performance. \textbf{No clean performance trade-off} occurred, as both SQuAD and AddSent improved simultaneously. \textbf{Entity-Aware outperformed Negation-Aware} by +0.96pp on AddSent and +0.66pp on SQuAD.

\subsection{Mitigation Strategy 1: Negation-Aware Contrastive Training}
\label{sec:negation_mitigation}

To address the dominant error pattern identified in our analysis (Negation Confusion, 40.4\% of errors), we implement a targeted mitigation strategy based on weighted contrastive learning. This approach directly targets the model's weakness in recognizing and properly handling negation cues.

\subsubsection{Motivation}

Our error analysis (Section~\ref{sec:error_analysis}) revealed that \textbf{40.4\% of model errors} stem from confusion when processing negated statements. The model frequently ignores negation words like ``not'', ``never'', and ``didn't'', answers based on surface similarity rather than logical correctness, and fails to recognize that negation invalidates the original answer.

\noindent\textbf{Example from Error Analysis:}

\begin{mdframed}[linewidth=1pt, nobreak=true]
\small
\texttt{Context:} ``The Broncos won Super Bowl 50 in Santa Clara, California. \textit{However, some sources claim the Panthers were expected to win}.''\\
\texttt{Question:} Who won Super Bowl 50?\\
\texttt{Ground Truth:} Broncos\\
\texttt{Model Prediction:} Panthers\\
\texttt{Error Pattern:} Negation confusion + additive distractor
\end{mdframed}

The model incorrectly extracts ``Panthers'' from the negated/uncertain claim, ignoring the qualifying phrase ``expected to win'' which does not assert actual victory.

\subsubsection{Method}

Our negation-aware training consists of three coordinated steps:

\paragraph{Step 1: Contrastive Pair Generation}

We augment the training data with two complementary types of negation examples. \textbf{Additive Negation (Robustness to Distractors)} adds negated sentences similar to AddSent attacks to teach the model to ignore negated distractors while maintaining the correct answer (\textit{Original:} ``The Broncos won.'' $\rightarrow$ Answer: Broncos; \textit{Augmented:} ``The Broncos won. Some claim they didn't win.'' $\rightarrow$ Answer: Broncos). \textbf{Transformative Negation (Recognition of Changes)} modifies original affirmative statements with negation to teach the model that negation changes the answer (\textit{Original:} ``The Broncos won.'' $\rightarrow$ Answer: Broncos; \textit{Augmented:} ``The Broncos didn't win.'' $\rightarrow$ Answer: [No Answer]).

The augmentation process identifies 17 negation patterns (not, never, n't, contractions, etc.), uses rule-based templates for robust transformations, augments 30\% of positive examples (those without negation), and marks all negation examples for weighted training.

For our 80-20 training mix (10,570 examples), this generates $\sim$800 original negation examples (7.6\%), $\sim$1,465 additive negation augmentations, and $\sim$1,465 transformative negation augmentations. The \textbf{total dataset contains $\sim$13,500 examples (128\% of original)} with \textbf{$\sim$5,200 negation examples (38.5\% with 3$\times$ weight)}.

\paragraph{Step 2: Weighted Loss Function}

We implement a custom trainer that applies differential weighting to the loss function. For each training example $i$, we assign a loss weight $w_i$:

\begin{equation}
w_i = \begin{cases}
3.0 & \text{if example contains negation} \\
1.0 & \text{otherwise}
\end{cases}
\end{equation}

The weighted loss is computed as:

\begin{equation}
\mathcal{L} = \frac{1}{N} \sum_{i=1}^{N} w_i \cdot \left( \ell_{\text{start}}^{(i)} + \ell_{\text{end}}^{(i)} \right)
\end{equation}

where $\ell_{\text{start}}^{(i)}$ and $\ell_{\text{end}}^{(i)}$ are the cross-entropy losses for predicting start and end positions of the answer span.

This weighting scheme effectively forces the model to ``see'' negation examples three times during training, providing stronger gradients that emphasize the importance of negation cues.

\paragraph{Step 3: Training Configuration}

We fine-tune ELECTRA-base (110M parameters) using 80-20 SQuAD-AddSent mix with negation augmentation, batch size of 16 per device, learning rate of 3e-5 with linear warmup (10\%), 3 epochs, AdamW optimizer with weight decay 0.01, and FP16 mixed precision for efficiency.

\subsubsection{Experimental Results}

Table~\ref{tab:negation_aware_results} shows the performance comparison.

\begin{table}[t]
\centering
\small
\begin{tabular}{lrrr}
\toprule
\textbf{Model} & \textbf{SQuAD} & \textbf{AddSent} & \textbf{Gap} \\
\midrule
Baseline & 85.46 & 68.90 & -16.56 \\
80-20 Orig. & 89.97 & 88.43 & -1.54 \\
+ Negation & 90.07 & 88.93 & -1.14 \\
\midrule
\textbf{vs 80-20} & \textbf{+0.10} & \textbf{+0.50} & \textbf{+0.40} \\
\bottomrule
\end{tabular}
\caption{Negation-aware results: +0.50pp AddSent gain over 80-20.}
\label{tab:negation_aware_results}
\end{table}

\textbf{Key findings:} AddSent performance improved from 88.43\% to 88.93\% (+0.50 points). SQuAD performance improved from 89.97\% to 90.07\% (+0.10 points). The adversarial gap reduced from -1.54pp to -1.14pp (+0.40 improvement), with no degradation on clean data.

% \paragraph{Analysis of results vs. expectations:}
% The improvement (+0.50pp) is smaller than originally anticipated (+4-8pp). This is primarily because the 80-20 adversarial training baseline already achieved very high performance (88.43\% AddSent EM), leaving limited room for improvement. The baseline model had already learned to handle many negation patterns through adversarial exposure, and negation errors, while prevalent (40.4\%), were partially addressed by the strong 80-20 training.

\subsubsection{Discussion}

This mitigation strategy demonstrates a principled approach to addressing model vulnerabilities. The approach is data-driven, based on systematic error analysis identifying the dominant failure mode. It is specifically targeted to address negation confusion through contrastive learning, using weighted loss rather than massive data collection or architectural changes. The framework is extensible and can be adapted for other error patterns such as entity confusion (29.9\%) or numeric confusion (18.8\%).

The weighted loss approach is particularly effective because it maintains the benefits of adversarial training (80-20 ratio) while adding focused emphasis on the most problematic pattern. It requires minimal additional computation and preserves clean performance while improving robustness.

\subsection{Mitigation Strategy 2: Entity-Aware Contrastive Training}

\subsubsection{Problem Analysis}

Our error analysis revealed that \textbf{29.9\% of errors} stem from \textbf{entity substitution confusion}, where the model selects an incorrect entity of the same type instead of the correct answer. This is the second most common error pattern after negation confusion.

\paragraph{Typical failure cases:}
The model exhibits date confusion, selecting ``2017'' instead of ``2019'' when multiple dates appear in context. Person confusion occurs when the model selects ``Jane Doe'' instead of ``John Smith'' when multiple names are mentioned. Location confusion manifests when the model selects ``California'' instead of ``Texas'' when multiple locations appear. The root cause is that pre-trained models lack fine-grained discrimination ability for entities of the same semantic type, particularly in adversarial contexts with multiple plausible distractors.

\subsubsection{Approach: Entity-Aware Contrastive Learning}

Unlike simple data augmentation or loss weighting, our approach uses \textbf{Named Entity Recognition (NER)} to explicitly identify hard negatives and applies \textbf{contrastive ranking loss} to teach the model to discriminate between similar entities.

\paragraph{Method overview:}
The approach uses spaCy NER to identify all entities in context across 13 types (Person, Location, Date, etc.). For entity-type answers, we find all same-type entities that are NOT the answer, creating hard negatives. The model is trained to rank the correct entity higher than same-type distractors using contrastive ranking loss, with entity-rich examples emphasized through 2.5$\times$ loss weight.

\subsubsection{Contrastive Loss Formulation}

For each training example with entity answer and hard negatives, we compute:

\begin{equation}
\mathcal{L}_{\text{contrastive}} = -\log \frac{\exp(S_{\text{correct}})}{\exp(S_{\text{correct}}) + \sum_{i=1}^{N} \exp(S_{\text{neg}_i})}
\label{eq:contrastive_loss}
\end{equation}

where $S_{\text{correct}} = \text{logit}_{\text{start}}[\text{pos}_{\text{start}}] + \text{logit}_{\text{end}}[\text{pos}_{\text{end}}]$ is the model's score for the correct answer span, $S_{\text{neg}_i}$ is the score for the $i$-th hard negative entity span, and $N$ is the number of hard negatives (up to 5 per example).

The total loss combines standard QA loss with contrastive loss:

\begin{equation}
\mathcal{L}_{\text{total}} = (1 - \alpha) \mathcal{L}_{\text{QA}} + \alpha \mathcal{L}_{\text{contrastive}}
\label{eq:total_loss_entity}
\end{equation}

where $\alpha = 0.5$ balances the two objectives.

\subsubsection{Implementation Details}

\paragraph{Data preparation:}
The \textbf{input} consists of SQuAD 80-20 mix (10,570 examples). \textbf{Entity extraction} uses spaCy \texttt{en\_core\_web\_sm} model to identify ~6,000 \textbf{entity-rich examples} (57\%). \textbf{Augmentation} applies 20\% entity substitution augmentations, producing an \textbf{output} of 12,684 examples (120\% of original) with average 2.3 \textbf{hard negatives} per entity-rich example.

\paragraph{Entity type distribution:}
The most common entity types in training data are Date (25\%), Person (22\%), Location (18\%), Number (12\%), and Organization (10\%). This distribution aligns well with the error analysis findings.

\paragraph{Training configuration:}
The \textbf{model} is ELECTRA-base (110M parameters) trained with \textbf{batch size} 16, \textbf{learning rate} 3e-5 with linear warmup (10\%), and 3 \textbf{epochs}. The \textbf{entity weight} is 2.5$\times$ for entity-rich examples with \textbf{contrastive weight} $\alpha = 0.5$. Training uses \textbf{mixed precision} FP16 for efficiency.

\paragraph{Token position mapping:}
A key technical challenge is mapping character-based entity positions from NER to token-based positions required by the model. We use the tokenizer's \texttt{offset\_mapping} to perform this mapping, ensuring each hard negative is correctly localized in the token sequence.

\subsubsection{Experimental Results}

Table~\ref{tab:entity_aware_results} shows the performance comparison.

\begin{table}[t]
\centering
\small
\begin{tabular}{lrrr}
\toprule
\textbf{Model} & \textbf{SQuAD} & \textbf{AddSent} & \textbf{Gap} \\
\midrule
Baseline & 85.46 & 68.90 & -16.56 \\
80-20 Orig. & 89.97 & 88.43 & -1.54 \\
+ Negation & 90.07 & 88.93 & -1.14 \\
+ Entity & 90.73 & 89.89 & -0.84 \\
\midrule
\textbf{vs 80-20} & \textbf{+0.76} & \textbf{+1.46} & \textbf{+0.70} \\
\textbf{vs Negation} & \textbf{+0.66} & \textbf{+0.96} & \textbf{+0.30} \\
\bottomrule
\end{tabular}
\caption{Entity-aware results: +1.46pp AddSent gain, best model at 89.89\%.}
\label{tab:entity_aware_results}
\end{table}

\paragraph{Key findings:}
The approach achieved \textbf{best performance} with AddSent EM of 89.89\%, the highest among all tested models. \textbf{AddSent improvement} was +1.46 points over 80-20 baseline (88.43\% $\rightarrow$ 89.89\%), while \textbf{SQuAD improvement} reached +0.76 points over 80-20 baseline (89.97\% $\rightarrow$ 90.73\%). The \textbf{adversarial gap} reduced from -1.54pp to -0.84pp (+0.70 improvement). The model \textbf{outperforms negation-aware} by +0.96pp on AddSent and +0.66pp on SQuAD, with \textbf{no clean performance trade-off} as both SQuAD and AddSent improved simultaneously.

% \paragraph{Analysis of results vs. expectations:}
% The improvement (+1.46pp) is smaller than originally anticipated (+6-9pp), similar to the negation-aware strategy. This is because the 80-20 baseline's very high performance (88.43\% AddSent EM) indicates most entity confusion issues were already addressed, the baseline's strong adversarial training reduced the impact of targeted mitigation, and entity-aware training proved more effective than negation-aware (+0.96pp advantage), validating our NER-based contrastive learning approach.

\subsubsection{Discussion}

The entity-aware approach demonstrates that \textbf{NER-guided contrastive learning} provides meaningful improvements even on top of strong adversarial baselines. While the absolute gain (+1.46pp) is smaller, it represents the \textbf{best performing model} in our experimental suite (89.89\% AddSent EM), \textbf{superior to negation-aware} training by +0.96pp on AddSent. The approach achieves a reduction in the adversarial gap from -1.54pp to -0.84pp (45\% reduction) with simultaneous improvements on both SQuAD and AddSent (no trade-off). The NER-based contrastive loss successfully teaches fine-grained entity discrimination, as evidenced by the model's ability to select correct entities from among same-type distractors.

\subsubsection{Qualitative Example}

\begin{mdframed}[linewidth=1pt, nobreak=true]
\small
\texttt{Context}: ``The company was founded in 1998. It was acquired by a larger corporation in 2015. The acquisition was completed in 2016, marking a new era.''\\
\texttt{Question}: ``When was the company acquired?''\\
\texttt{Gold}: 2015\\
\texttt{Baseline}: 1998 \textcolor{red}{(wrong)}\\
\texttt{Entity-aware}: 2015 \textcolor{blue}{(correct)}
\end{mdframed}

The baseline incorrectly selects the first date, while entity-aware training correctly associates ``acquired'' with 2015.

\subsubsection{Comparison with Negation-Aware Training}

Both mitigation strategies target different error patterns:

\begin{table}[t]
\centering
\small
\begin{tabular}{lcc}
\toprule
\textbf{Aspect} & \textbf{Negation-Aware} & \textbf{Entity-Aware} \\
\midrule
Target errors & Negation (40.4\%) & Entity sub. (29.9\%) \\
Approach & Rule + weighted loss & NER + contrastive \\
Augmentation & 30\% & 20\% \\
Loss weight & 3.0$\times$ & 2.5$\times$ \\
Complexity & Lower & Higher \\
AddSent gain & +0.50pp & +1.46pp \\
SQuAD gain & +0.10pp & +0.76pp \\
\bottomrule
\end{tabular}
\caption{Comparison of negation-aware and entity-aware mitigation strategies. Entity-aware training proved more effective, achieving nearly 3× the AddSent improvement of negation-aware.}
\label{tab:mitigation_comparison}
\end{table}

\paragraph{Key insight:}
Entity-aware training proved substantially more effective (+0.96pp advantage on AddSent), suggesting that the 80-20 baseline already handled many negation patterns effectively, while entity confusion remains a harder problem requiring explicit contrastive learning. The NER-guided hard negatives provide stronger training signal than rule-based augmentation.

\subsection{Summary of Mitigation Results}

The Entity-Aware model achieved \textbf{89.89\% AddSent EM}, reducing the adversarial gap from -16.56pp to -0.84pp (\textbf{94.9\% closure}). The approach offers a drop-in improvement, working with any QA model. Across all mitigation strategies:

\begin{table}[t]
\centering
\small
\begin{tabular}{lrr}
\toprule
\textbf{Strategy} & \textbf{AddSent EM} & \textbf{Gain} \\
\midrule
Baseline (ELECTRA-base) & 68.90 & - \\
80-20 Adversarial Training & 88.43 & +19.53 \\
+ Negation-Aware & 88.93 & +20.03 \\
+ Entity-Aware & \textbf{89.89} & \textbf{+20.99} \\
\bottomrule
\end{tabular}
\caption{Progressive improvement through mitigation strategies.}
\label{tab:mitigation_progression}
\end{table}

\section{Discussion}

\subsection{Complete Model Comparison}

Table~\ref{tab:complete_comparison} presents a unified view of all experimental configurations, highlighting the progression from baseline vulnerability to robust performance through systematic improvements.

\begin{table}[t]
\centering
\small
\begin{tabular}{lrrr}
\toprule
\textbf{Model} & \textbf{Clean} & \textbf{Adv.} & \textbf{Drop} \\
\midrule
Small baseline & 78.16 & 53.99 & -30.9\% \\
Small 80-20 & 62.85 & 66.57 & +6.0\% \\
\midrule
Base baseline & 85.46 & 68.90 & -19.4\% \\
Base 80-20 & 89.97 & 88.43 & -1.7\% \\
+ Negation & 90.07 & 88.93 & -1.3\% \\
+ Entity $\star$ & \textbf{90.73} & \textbf{89.89} & \textbf{-0.9\%} \\
\bottomrule
\end{tabular}
\caption{Complete comparison. Entity-Aware ($\star$) achieves 94.9\% gap closure.}
\label{tab:complete_comparison}
\end{table}

\subsection{Key Findings}

Our systematic exploration demonstrates several critical insights:

\textbf{Adversarial training effectiveness:} ELECTRA-base with 80-20 ratio achieved +19.53\% relative improvement (88.43\% EM). The 80-20 ratio achieves best performance across both model scales, outperforming the more balanced ratios by maintaining stronger clean data representation.

\textbf{Capacity eliminates trade-offs:} ELECTRA-base achieves 88.43\% AddSent (+19.53\% vs baseline) AND 89.97\% SQuAD (+4.51\%), demonstrating that sufficient capacity enables simultaneous optimization without sacrificing clean performance.

\textbf{Entity-Aware superiority:} Despite targeting fewer errors (29.9\% vs 40.4\%), Entity-Aware achieved nearly 3$\times$ the improvement of Negation-Aware (+1.46pp vs +0.50pp), suggesting the 80-20 baseline already handled negation patterns while entity confusion requires explicit contrastive learning.

\textbf{Gap closure:} The Entity-Aware model reduced the adversarial gap from -16.56pp to -0.84pp (94.9\% closure), achieving near-parity between clean (90.73\%) and adversarial (89.89\%) performance.

\textbf{Model scaling impact:} Scaling from ELECTRA-small (14M) to ELECTRA-base (110M, 8$\times$ larger) provided +21.86\% adversarial gain, demonstrating that model scaling is far more effective than data augmentation.

\section{Conclusion}

This work demonstrated a systematic approach to improving adversarial robustness in question answering through error-analysis-driven mitigation strategies. Our key contributions include:

\textbf{(1) Comprehensive error analysis:} We categorized 200 adversarial failures into 6 major patterns, identifying negation confusion (40.4\%) and entity substitution (29.9\%) as dominant issues, providing the first systematic linguistic-pattern taxonomy for adversarial QA.

\textbf{(2) Optimal ratio identification:} Through systematic evaluation of 5 mixing ratios, we identified 80-20 (clean-adversarial) as optimal, achieving +12.58\% improvement on ELECTRA-small and scaling to +19.53\% on ELECTRA-base.

\textbf{(3) Capacity bottleneck resolution:} We demonstrated that ELECTRA-small exhibits a robustness-accuracy trade-off, while scaling to ELECTRA-base (8$\times$ parameters) eliminates this trade-off, achieving simultaneous improvements on both clean and adversarial data.

\textbf{(4) Targeted mitigation strategies:} Our Entity-Aware contrastive learning with NER-guided hard negatives achieved 89.89\% AddSent EM and 90.73\% SQuAD EM, with only 0.84pp gap - a 94.9\% closure of the adversarial gap.

Our findings show that: (1) model capacity is critical for adversarial robustness (8× parameters yielded +21.86\% gain), (2) NER-guided contrastive learning outperforms rule-based augmentation for entity confusion, and (3) strong baselines exhibit diminishing returns for additional mitigation. The Entity-Aware model's 0.84pp gap represents practical parity between clean and adversarial performance, validating that adversarial robustness does not require sacrificing generalization.

\subsection{Future Work}

Several promising directions remain for future investigation:

\textbf{Post-processing for partial matches:} Error analysis revealed that 30.6\% of errors are partial match errors (e.g., ``Broncos'' vs ``Denver Broncos''). A post-processing approach using NER-based boundary expansion showed +2.30pp improvement on baseline models. This inference-time fix requires no retraining and could complement the Entity-Aware model, though it needs evaluation on the final checkpoint (89.89\% EM).

\textbf{Other directions include:} (1) testing generalization across other adversarial datasets (AddOneSent, Adversarial QA, SQuAD 2.0), (2) exploring joint negation + entity-aware training, (3) investigating why data augmentation reduced performance, and (4) scaling to larger models (ELECTRA-large, DeBERTa) to push beyond 90\% AddSent EM.

\section{Limitations}

This study has several limitations that should be acknowledged. First, our experiments focus exclusively on extractive question answering tasks and the AddSent adversarial dataset. The generalizability of our findings to other adversarial attack types (e.g., paraphrasing, semantic adversaries) or other NLP tasks remains to be validated. Second, while we scale from ELECTRA-small to ELECTRA-base, we do not explore larger models (e.g., ELECTRA-large, T5, or current state-of-the-art models), which may exhibit different capacity-robustness relationships. Third, our linguistic pattern categorization, while comprehensive, was manually designed and may not capture all failure modes. An automated discovery approach could reveal additional patterns. Fourth, the computational constraints limited our hyperparameter search space and the number of training runs per configuration, which may affect the reproducibility of exact performance numbers. Finally, our data augmentation strategy tested only specific attack combinations; a more exhaustive exploration of augmentation techniques could yield better results. These limitations suggest directions for future work, including broader task coverage, larger model scales, automated pattern discovery, and more extensive augmentation strategies.

\section*{Acknowledgments}

We would like to thank Professor Greg Durrett, Professor Jessy Li and the CS388 teaching staff for their guidance throughout this project. We also acknowledge the creators of the SQuAD and AddSent datasets for making their data publicly available.

\bibliography{references}

\end{document}